\documentclass[runningheads]{llncs}
\usepackage[T1]{fontenc}
\usepackage{lmodern}
\usepackage{graphicx}
\usepackage{amsmath, amssymb}
\usepackage{booktabs}
\usepackage{multirow}
\usepackage{float}
\usepackage{color}

\usepackage{array}

\usepackage{algorithm}
\usepackage{algpseudocode}
\usepackage{hyperref}

\hypersetup{
    colorlinks=true,
    linkcolor=blue,
    citecolor=blue,
    urlcolor=blue
}

\title{DDEL: Closed-Loop Mutual Enhancement of Dataset Distillation and Ensemble Learning}

\titlerunning{DDEL}
\author{Weiqi Zhang\inst{1}\thanks{Email: \href{mailto:weiqizhang@chiba-u.jp}{\nolinkurl{weiqizhang@chiba-u.jp}}.} \and Yongquan Yang\inst{2}\thanks{Email: \href{mailto:remy_yang@foxmail.com}{\nolinkurl{remy_yang@foxmail.com}} or \href{mailto:yongquan.yang@sciences4ai.com}{\nolinkurl{yongquan.yang@sciences4ai.com}}.}}
\authorrunning{W. Zhang et al.}
\institute{Independent Researcher, Chengdu, China \and Institute of Sciences for AI, Chengdu, Sichuan, China}

\begin{document}
\maketitle

\begin{abstract}
Dataset distillation (DD) reduces a training set to a small set of examples, while ensemble learning (EL) combines predictions from several models. When applied in sequence, the distilled set remains fixed as the ensemble learns. We introduce DDEL, a framework that feeds ensemble behavior back into distillation. Each round trains the ensemble, measures loss and prediction disagreement on the source data, and uses these signals to guide the next data update. We implement this loop with distribution matching and a homogeneous three-member ensemble.

Preliminary experiments cover skin lesion classification (ISIC 2019), treatment-response prediction (I-SPY1), and binary two-year prognosis classification (NSCLC-Radiomics). With the same images-per-class budget as the open-loop comparison, DDEL reaches balanced accuracy $0.3471$ on ISIC and mean ROC-AUC $0.5120$ and $0.5597$ on I-SPY1 and NSCLC. The I-SPY1 difference is small, and the ISIC trajectory shows that more rounds do not consistently improve performance. These results offer initial evidence for studying DD--EL feedback, with its computational cost and robustness still requiring controlled evaluation.


\end{abstract}

\section{Introduction}
Medical image models often require substantial training data and computation \cite{litjens2017survey,zhou2021reviewmedai}. Dataset distillation (DD) aims to reduce the data needed for downstream training by learning a small training set \cite{wang2018datasetdistillation,zhao2021gradientmatching,cazenavette2022datasetdistill}; medical applications also explore compressed data sharing \cite{li2022datasetmedical,li2022compressedgastric}. Ensemble learning (EL) addresses a different problem: combining models can improve prediction and provide information about uncertainty \cite{lakshminarayanan2017deepensemble,ovadia2019uncertainty,yang2022ensemble_survey}. A straightforward way to combine the two is to distill once and then train an ensemble. In this open-loop setup, what the ensemble learns about the source samples cannot affect the distilled set.

Ensemble behavior provides a possible basis for revising that set. Average member loss identifies difficult samples, while prediction disagreement identifies samples on which members differ. These signals can guide which source examples distillation emphasizes. The revised data then change what the ensemble sees in its next training round. Yang proposed this connection between DD and EL \cite{yang2025mutual}. Here, we give it an explicit update rule and examine one implementation in medical imaging.

DDEL (Dataset Distillation + Ensemble Learning) alternates three steps: train the ensemble on the current distilled set, evaluate its members on a fixed source pool, and update the distilled set using the resulting signals. We use distribution matching (DM) \cite{zhao2023distributionmatching} and introduce class-wise source-sampling weights based on normalized loss and disagreement. Our experiments use three members of the same architecture: ResNet-50 \cite{he2016resnet} on ISIC and Res3D-18 \cite{tran2018closerlook} on I-SPY1 and NSCLC.

We compare DDEL with single-model training, full-data ensembles, DD-only, and ensembles trained on fixed distilled data (EL-on-DD). The three DD-based settings use the same distilled-set size within each task. The experiments cover skin lesion classification, treatment response, and binary two-year prognosis; ISIC results also include calibration metrics. This report contributes the feedback formulation, a DM implementation, and preliminary comparisons. Its aim is to document the approach and observations rather than establish a general advantage across distillation methods.

\section{Related Work}
\label{sec:related}

\subsection{Dataset Distillation}
DD aims to learn a compact training set that retains useful information for downstream learning \cite{geng2023surveydd}. Gradient matching aligns gradients computed from real and synthetic samples \cite{zhao2021gradientmatching}. Trajectory matching matches network parameters reached after multiple training steps \cite{cazenavette2022datasetdistill}. Distribution matching aligns real and synthetic feature statistics \cite{zhao2023distributionmatching}. Other approaches use pretrained generative priors \cite{cazenavette2023generativeprior}, and differentiable augmentation can be used during condensation \cite{zhao2021siamese}. Medical studies have explored DD for compressed data sharing \cite{li2022datasetmedical,li2022compressedgastric}.

These methods primarily address how to construct the compact set. Our question concerns the next stage: after models have learned from that set, can their behavior on the source data help revise it? DDEL uses DM for this update, while keeping the feedback interface separate from the choice of distillation method.

\subsection{Ensemble Learning and Feedback}
Ensemble methods combine predictions from several models \cite{yang2022ensemble_survey}. Deep ensembles average predicted probabilities for classification and have been studied for predictive uncertainty estimation \cite{lakshminarayanan2017deepensemble}. The quality of uncertainty estimates can change under dataset shift \cite{ovadia2019uncertainty}. Neural-network calibration has been studied in general classification settings \cite{guo2017calibration}; calibration and uncertainty have also been studied in medical image analysis \cite{mehrtash2020confidence,lambert2024trustworthy}. Training several models introduces an additional computational cost \cite{yang2022ensemble_survey}.

In DDEL, member loss and disagreement are used during data refinement, beyond their usual role in prediction analysis. Following the conceptual proposal of Yang \cite{yang2025mutual}, we connect these signals to the next distillation round. An ensemble trained on fixed distilled data provides the open-loop reference for our experiments.

\section{Method}
\label{sec:method}

\subsection{Overview}

DDEL alternates ensemble training and distilled-data updates. The source data remain fixed throughout the loop.

We model training with two states:
\begin{itemize}
\item \textbf{Distilled dataset state} $\Sigma_{\mathrm{dm}}^{(t)}$: effective training distribution at iteration $t$;
\item \textbf{Ensemble state} $\Theta^{(t)}=\{\theta_m^{(t)}\}_{m=1}^M$: parameters of $M$ ensemble members.
\end{itemize}

Each round applies the ensemble update $M$, signal extraction $W$, and data update $S_{\mathrm{DM}}$ in that order:
\begin{equation}
\begin{aligned}
\Theta^{(t+1)} &= M(\Sigma_{\mathrm{dm}}^{(t)}, \Theta^{(t)}),\\
\mathbf{q}_{\mathrm{src}}^{(t)} &= W(\Theta^{(t+1)}, \Sigma_{\mathrm{src}}),\\
\Sigma_{\mathrm{dm}}^{(t+1)} &= S_{\mathrm{DM}}(\mathbf{q}_{\mathrm{src}}^{(t)}, \Sigma_{\mathrm{src}}).
\end{aligned}
\label{eq:wsm}
\end{equation}

\begin{figure}[t]
\centering
\includegraphics[width=\linewidth]{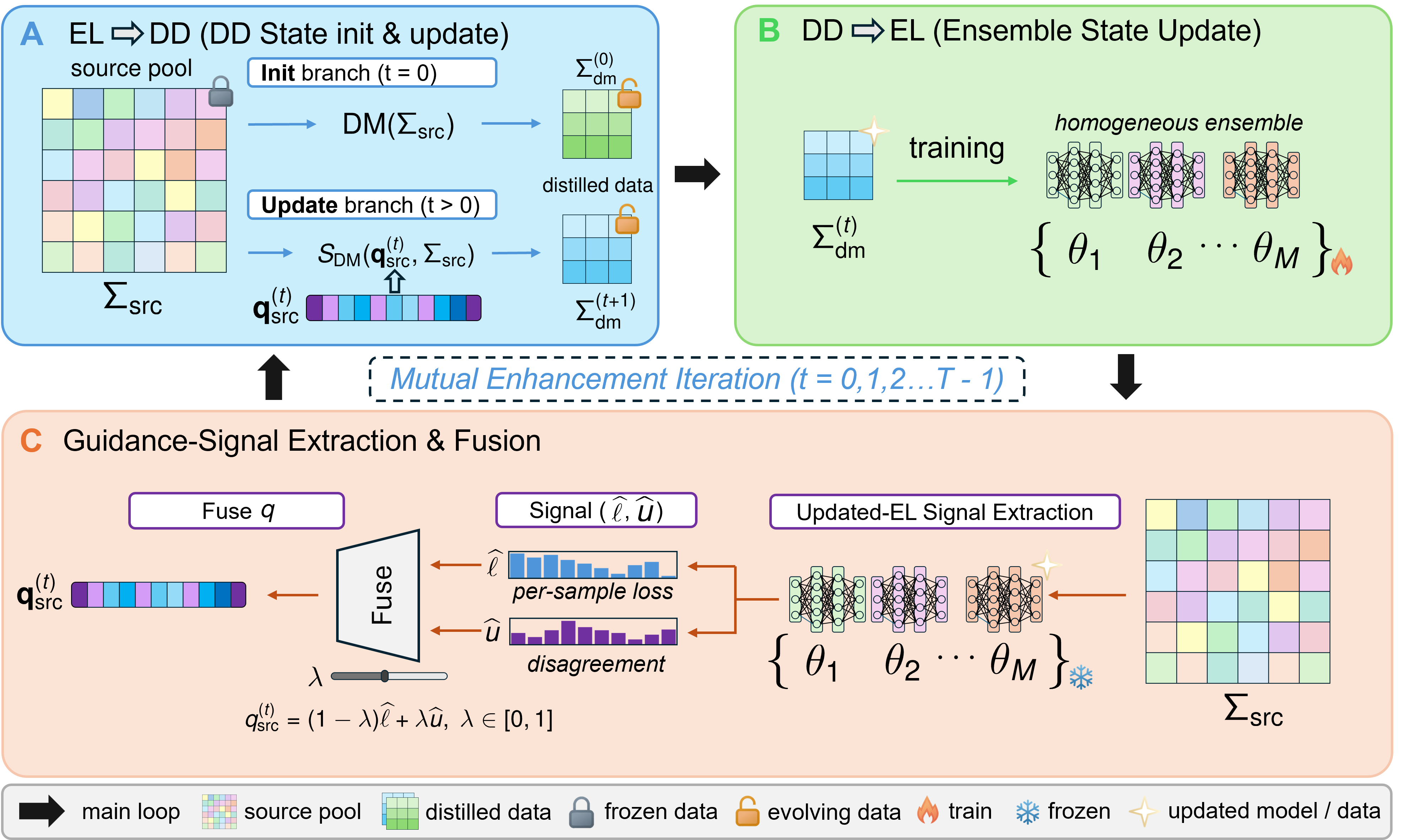}
\caption{DDEL feedback loop. \textbf{(A)} DM initializes the distilled set from the fixed source pool; subsequent updates use ensemble-derived signals. \textbf{(B)} The ensemble is trained on the current distilled set. \textbf{(C)} Frozen members predict on the source pool, producing loss and disagreement signals for the next data update. Initialization precedes the loop; the update branch is used after each training round, including $t=0$ in Algorithm~\ref{alg:ddel}.}

\label{fig:ddel_overview}
\end{figure}

This yields the closed-loop transition (one outer iteration)
\[
\Sigma_{\mathrm{dm}}^{(t)} \rightarrow \Theta^{(t+1)} \rightarrow \mathbf{q}_{\mathrm{src}}^{(t)}
\rightarrow \Sigma_{\mathrm{dm}}^{(t+1)},
\]
The fixed source pool $\Sigma_{\mathrm{src}}$ is used for signal extraction and data synthesis. Figure~\ref{fig:ddel_overview} and Algorithm~\ref{alg:ddel} show the same sequence.

\subsection{State Initialization}

Let $\mathcal{D}=\{(x_i,y_i)\}_{i=1}^{N}$ denote the training source pool. We initialize
\[
\Sigma_{\mathrm{src}} \equiv \mathcal{D},\quad \Sigma_{\mathrm{dm}}^{(0)} \leftarrow \mathrm{DM}(\Sigma_{\mathrm{src}}),
\]
Here, $\mathrm{DM}(\cdot)$ denotes distribution-matching distillation \cite{zhao2023distributionmatching}. Ensemble parameters $\Theta^{(0)}$ are initialized according to the task setup.

\subsection{Ensemble-Induced Signal Operator $W(\cdot)$}

After training round $t$, $W$ evaluates the members of $\Theta^{(t+1)}$ on $\Sigma_{\mathrm{src}}$ without updating their parameters.

For sample $x_i$, let $p_{i,m}=f_{\theta_m}(x_i)$ denote the class-probability vector of member $m$, and let the ensemble mean be
\[
\bar{p}_i=\frac{1}{M}\sum_{m=1}^{M}p_{i,m}.
\]

We compute two signals for each sample. With $\ell$ denoting classification cross-entropy, the loss signal is the mean of the member losses:

\paragraph{Loss signal}
\begin{equation}
\ell_i = \frac{1}{M}\sum_{m=1}^{M}\ell\!\left(f_{\theta_m}(x_i), y_i\right).
\label{eq:loss_signal}
\end{equation}

\paragraph{Disagreement signal}
\begin{equation}
u_i=\frac{1}{M}\sum_{m=1}^{M}\left\|p_{i,m}-\bar{p}_i\right\|_2^2.
\label{eq:dis_signal}
\end{equation}

We normalize the signals and combine them using
\begin{equation}
q_{\mathrm{src},i}^{(t)} = (1-\lambda)\,\widehat{\ell}_i + \lambda\,\widehat{u}_i,\quad \lambda\in[0,1],
\label{eq:fuse_q}
\end{equation}
where $\widehat{\ell}_i$ and $\widehat{u}_i$ are the normalized loss and disagreement. The loss in Eq.~\eqref{eq:loss_signal} differs from cross-entropy computed after averaging predictions.
The vector $\mathbf{q}_{\mathrm{src}}^{(t)}$ supplies the sampling signal for the data update.

\subsection{Dataset Update Operator $S_{\mathrm{DM}}(\cdot)$}

The data update maps source signals to a new distilled set:
\[
\Sigma_{\mathrm{dm}}^{(t+1)}=S_{\mathrm{DM}}(\mathbf{q}_{\mathrm{src}}^{(t)},\Sigma_{\mathrm{src}}).
\]

\paragraph{DM implementation.}
We build on class-wise DM \cite{zhao2023distributionmatching} and introduce feedback-based source sampling. Within each class, our signals determine sampling probabilities
\begin{equation}
w_i^{(t)} = \frac{\exp\!\left(\alpha q_{\mathrm{src},i}^{(t)}\right)}{\sum_{j:y_j=y_i}\exp\!\left(\alpha q_{\mathrm{src},j}^{(t)}\right)}, \quad \alpha>0,
\label{eq:weight_dm}
\end{equation}
These probabilities guide real-sample selection for DM initialization and class-wise matching batches. The resulting synthetic set defines $\Sigma_{\mathrm{dm}}^{(t+1)}$.

\paragraph{Other data updates.}
Other compact-data constructions include coreset selection \cite{mirzasoleiman2020coresets}, gradient matching \cite{zhao2021gradientmatching}, and trajectory matching \cite{cazenavette2022datasetdistill}. Adapting these methods to the feedback interface is a possible extension; their compatibility has not been evaluated. We evaluate only the DM implementation.

\subsection{Ensemble Update Operator $M(\cdot)$}

At outer iteration $t$, each member is trained on the current synthesized distilled set $\Sigma_{\mathrm{dm}}^{(t)}$:
\begin{equation}
\mathcal{L}_{\mathrm{dm}}^{(t)}(\Theta)=\frac{1}{M|\Sigma_{\mathrm{dm}}^{(t)}|}\sum_{m=1}^{M}\sum_{(x,y)\in \Sigma_{\mathrm{dm}}^{(t)}}\ell\!\left(f_{\theta_m}(x),y\right).
\end{equation}
The corresponding parameter update is
\[
\Theta^{(t+1)} \leftarrow \mathrm{Update}_M\!\big(\Theta^{(t)};\Sigma_{\mathrm{dm}}^{(t)}\big),
\]
After the prescribed optimization steps on $\Sigma_{\mathrm{dm}}^{(t)}$, the updated parameters are passed to $W$.

\subsection{Feedback Assumptions}
Loss and disagreement determine which source samples receive more weight. High values may reflect useful difficult examples, but they can also arise from noise or incorrect labels. The loop therefore has no guarantee of improving generalization, calibration, or a common objective at every round.

\subsection{Closed-Loop Training Algorithm}

Algorithm~\ref{alg:ddel} summarizes DDEL.

\begin{algorithm}[t]
\caption{DDEL (minimal instantiation)}
\label{alg:ddel}
\begin{algorithmic}[1]
\Require frozen source pool $\Sigma_{\mathrm{src}}$, ensemble size $M$, outer iterations $T\ge 1$
    \State \textbf{Initialize:} $\Sigma_{\mathrm{dm}}^{(0)} \leftarrow \mathrm{DM}(\Sigma_{\mathrm{src}})$; initialize $\Theta^{(0)}$ (e.g., random)
\For{$t = 0,1,\ldots,T-1$}
    \State $\Theta^{(t+1)} \leftarrow M(\Sigma_{\mathrm{dm}}^{(t)},\Theta^{(t)})$
    \State $\mathbf{q}_{\mathrm{src}}^{(t)} \leftarrow W(\Theta^{(t+1)},\Sigma_{\mathrm{src}})$
    \State $\Sigma_{\mathrm{dm}}^{(t+1)} \leftarrow S_{\mathrm{DM}}(\mathbf{q}_{\mathrm{src}}^{(t)},\Sigma_{\mathrm{src}})$
\EndFor
\State \Return $\Theta^{(T)}$
\end{algorithmic}
\end{algorithm}

The returned model is $\Theta^{(T)}$, trained on $\Sigma_{\mathrm{dm}}^{(T-1)}$ in the last round. The final data update produces $\Sigma_{\mathrm{dm}}^{(T)}$, which is available for further training but is not used by the returned model. Thus, with $T=1$, feedback has not yet affected model training; its effect starts in the second round.

\subsection{Scope}
The operator definitions describe the feedback interface. The experiments below concern one implementation: DM guided by loss and disagreement from a homogeneous ensemble. They cannot establish benefits for other distillation methods or separate feedback effects from those of additional optimization.

\section{Experiments}
\label{sec:exp}

\subsection{Experimental Setup}

\paragraph{Tasks and datasets.}
We study skin lesion classification on ISIC 2019 \cite{isic2019challenge,tschandl2018ham10000,codella2017isic,hernandezperez2024bcn20000}, binary treatment-response prediction on I-SPY1 \cite{ispy1tcia}, and binary two-year prognosis classification on NSCLC-Radiomics \cite{nsclc_radiomics_tcia}. For NSCLC, the two-year outcome is treated as a binary classification target.

\paragraph{Protocol and reporting.}
ISIC uses the reported official split and images per class (IPC)=5. I-SPY1 and NSCLC use stratified five-fold cross-validation and IPC=4. We use ResNet-50 \cite{he2016resnet} for ISIC and Res3D-18 \cite{tran2018closerlook} for I-SPY1/NSCLC. We use a fixed reporting round, $t=2$, for the main DDEL comparison. The ISIC trajectory also includes rounds through $t=4$. ISIC entries are single-split values; binary-task entries are fold-wise mean$\pm$standard deviation. No significance tests or cross-seed stability estimates are available.

\paragraph{Comparison settings.}
Baseline trains a single model on full data, and EL-only trains a full-data ensemble. DD-only trains one model on a DM set \cite{zhao2023distributionmatching}; EL-on-DD trains an ensemble on that fixed set. DDEL adds repeated signal extraction and data updates. The three DD-based settings use the same IPC within each task. Their total computation differs, since DDEL repeats training and synthesis. The full-data settings provide performance context.

\paragraph{Metrics.}
We report balanced accuracy (macro-averaged recall) for ISIC and ROC-AUC for the binary tasks. For ISIC, we also report Brier score \cite{brier1950verification} and negative log-likelihood (NLL) \cite{guo2017calibration} for probability quality, and expected calibration error (ECE) \cite{guo2017calibration} for calibration. Calibration results for the binary tasks and systematic mechanism ablations are left for further work.

\subsection{Results}

\begin{table}[htbp]
\centering
\caption{Discrimination at matched IPC for the DD-based methods, with full-data references below. Bold marks the highest value within the DD-based group. DDEL uses round $t=2$. ISIC values are from one split; binary-task values are fold-wise mean$\pm$standard deviation.}
\label{tab:main_results}
\small
\setlength{\tabcolsep}{4pt}
\begin{tabular}{lccc}
\toprule
Method & ISIC 2019 & I-SPY1 & NSCLC \\
 & Bal.Acc $\uparrow$ & AUC $\uparrow$ & AUC $\uparrow$ \\
 & IPC=5 & IPC=4 & IPC=4 \\
\midrule
DD-only & 0.2798 & $0.4867\pm0.0691$ & $0.5025\pm0.0528$ \\
EL-on-DD & 0.3056 & $0.5081\pm0.0802$ & $0.5094\pm0.0279$ \\
DDEL ($t=2$) & \textbf{0.3471} & $\mathbf{0.5120}\pm0.0452$ & $\mathbf{0.5597}\pm0.0420$ \\
\midrule
\multicolumn{4}{l}{\textit{Full-data references}} \\
Baseline (single) & 0.5703 & $0.4706\pm0.0417$ & $0.5509\pm0.0387$ \\
EL-only & 0.5900 & $0.5210\pm0.0445$ & $0.5536\pm0.0677$ \\
\bottomrule
\end{tabular}
\end{table}

Table~\ref{tab:main_results} compares DDEL with fixed-data EL-on-DD at the same IPC. This comparison includes both feedback and additional optimization; it does not isolate their contributions.

\paragraph{ISIC 2019 (official split, IPC=5).}
Balanced accuracy gives equal weight to each class despite the uneven class frequencies. At $t=2$, DDEL reaches $0.3471$, compared with $0.2798$ for DD-only and $0.3056$ for EL-on-DD: absolute increases of $0.0673$ and $0.0415$. A substantial gap remains to the full-data references.

\begin{table}[htbp]
\centering
\caption{ISIC test metrics across DDEL rounds on one split. Bold and underlining mark the best and second-best values in each column. The main comparison uses $t=2$.}
\label{tab:isic_ddel_per_t}
\begingroup\small
\begin{tabular}{c|ccccc}
\hline
$t$ & Acc $\uparrow$ & Bal.Acc $\uparrow$ & Brier $\downarrow$ & ECE $\downarrow$ & NLL $\downarrow$ \\
\hline
0 & 0.2896 & 0.2948 & 0.8504 & \underline{0.1285} & 1.9831 \\
1 & \textbf{0.3701} & \textbf{0.3582} & \textbf{0.7658} & \textbf{0.1020} & \textbf{1.8222} \\
2 & 0.3059 & 0.3471 & 0.8629 & 0.1838 & 1.9819 \\
3 & \underline{0.3615} & 0.3463 & \underline{0.8171} & 0.1420 & \underline{1.8309} \\
4 & 0.2638 & \underline{0.3542} & 0.8792 & 0.1560 & 2.0563 \\
\hline
\end{tabular}
\endgroup
\end{table}

All five metrics are best at $t=1$ in Table~\ref{tab:isic_ddel_per_t}. Later rounds fluctuate, so additional feedback does not consistently improve performance. We retain $t=2$ for the fixed-round main comparison.

\begin{table}[htbp]
\centering
\caption{ISIC probabilistic prediction quality and calibration on one split. Bold marks the best value within the DD-based group.}
\label{tab:isic_calib}
\small
\begin{tabular}{lccc}
\toprule
Method & Brier $\downarrow$ & ECE $\downarrow$ & NLL $\downarrow$ \\
\midrule
DD-only & 1.0042 & 0.3076 & 2.4509 \\
EL-on-DD & 0.8671 & \textbf{0.1506} & 2.0207 \\
DDEL ($t=2$) & \textbf{0.8629} & 0.1838 & \textbf{1.9819} \\
\midrule
\multicolumn{4}{l}{\textit{Full-data references}} \\
Baseline (single) & 0.4802 & 0.1756 & 1.2278 \\
EL-only & 0.3868 & 0.0657 & 0.8674 \\
\bottomrule
\end{tabular}
\end{table}

\paragraph{ISIC probability quality and calibration.}
At $t=2$, DDEL has slightly lower Brier score and NLL than EL-on-DD, while its ECE is higher (Table~\ref{tab:isic_calib}).

\paragraph{I-SPY1 (five-fold CV, IPC=4).}
DDEL obtains mean ROC-AUC $0.5120$, compared with $0.4867$ for DD-only and $0.5081$ for EL-on-DD. Its advantage over EL-on-DD is only $0.0039$, with AUC close to chance. The smaller fold-wise standard deviation alone cannot establish a reliable improvement or stability across seeds.

\paragraph{NSCLC-Radiomics (five-fold CV, IPC=4).}
DDEL obtains mean ROC-AUC $0.5597$, compared with $0.5025$ for DD-only and $0.5094$ for EL-on-DD. The increase over EL-on-DD is $0.0503$, although its statistical significance has not been assessed. The full-data references have means $0.5509$ and $0.5536$. Similar means under these different training budgets do not establish equivalence or superiority.

\section{Discussion}
\label{sec:discuss}
DDEL makes ensemble behavior an input to data refinement. In this DM implementation, all three tasks have higher reported discrimination values than EL-on-DD at the same IPC. The magnitude varies: I-SPY1 changes little, while ISIC and NSCLC show larger differences. ISIC also shows fluctuating round-wise performance and mixed probability-quality and calibration results. These observations favor treating feedback as a choice to evaluate for each task, rather than assuming that repeated updates will help.

The main comparison combines feedback with extra optimization. Controls with matched computation and signal ablations are needed to distinguish their effects. The ISIC findings also require replication across splits and seeds, while the binary-task results need independent seed runs and external testing. Results from this implementation provide no evidence for other DD operators.

Exact reproduction requires further documentation of normalization, distillation settings, optimizer-state handling, and timing definitions. The source pool must contain training samples only. Patient grouping, MRI time points, and treatment of censored two-year outcomes also need to be specified. Binary prognosis classification leaves time-to-event information unused. Broader calibration studies and clinically justified decision thresholds would be needed before interpreting these predictions as clinical risk estimates.

\section{Conclusion}
DDEL connects ensemble training to distillation through sample-level loss and disagreement. Preliminary DM experiments show higher discrimination values than fixed-data EL-on-DD at the same IPC, with a small difference on I-SPY1 and mixed probability-quality, calibration, and iteration effects on ISIC. The framework offers a way to study this feedback, while further controlled experiments are needed to determine when its benefits justify the extra computation.

\paragraph{AI assistance.}
OpenAI Codex was used for language editing, formatting, and manuscript consistency checks. The authors take responsibility for the final manuscript.


\bibliographystyle{splncs04}
\bibliography{references}

\end{document}